\documentclass[letterpaper]{article} 
\usepackage{aaai24}  
\usepackage{times}  
\usepackage{helvet}  
\usepackage{courier}  
\usepackage[hyphens]{url}  
\usepackage{graphicx} 
\usepackage{natbib}  
\usepackage{caption} 
\usepackage{algorithm}
\usepackage{algorithmic}
\usepackage{newfloat}
\usepackage{listings}
\DeclareCaptionStyle{ruled}{labelfont=normalfont,labelsep=colon,strut=off} 
\floatstyle{ruled}
\newfloat{listing}{tb}{lst}{}
\floatname{listing}{Listing}
\usepackage{amsmath}
\usepackage{amsfonts} 
\usepackage{booktabs} 
\newcommand{\eg}{\emph{e.g.,}~}

\title{ConditionVideo: Training-Free Condition-Guided Video Generation}
\author{
    Bo Peng$^{1,2}\thanks{Work done as an intern at Shanghai AI Lab.}$, Xinyuan Chen$^{2}\thanks{Corresponding author}$, Yaohui Wang$^{2}$, Chaochao Lu$^{2}$, Yu Qiao$^{2}$
}
\affiliations{
    $^1$Shanghai Jiao Tong University \\ 
    $^2$Shanghai Artificial Intelligence Laboratory\\
    \{pengbo,chenxinyuan,wangyaohui,luchaochao,qiaoyu\}@pjlab.org.cn 

}

\usepackage{bibentry}

\begin{document}
\maketitle
\begin{figure*}[t]
\centering
\includegraphics[width=0.85\textwidth]{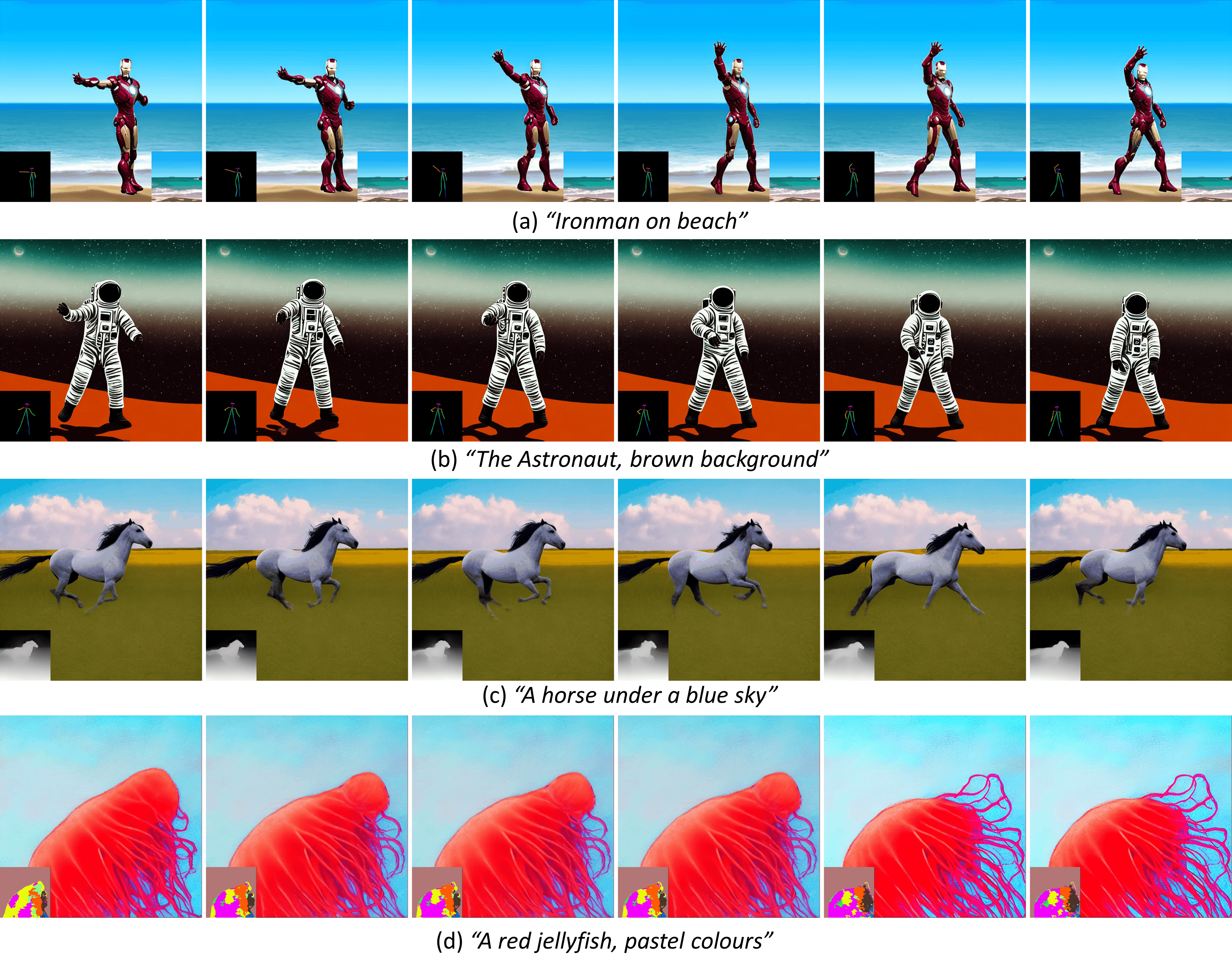}
\captionof{figure}{Our training-free method generates videos conditioned on different inputs. In (a), the illustration showcases the process of generation using provided scene videos and pose information, with the background wave exhibiting a convincingly lifelike motion. (b), (c), and (d) are generated based on condition only, which are pose, depth, and segmentation, respectively.}
\label{fig:result}
\end{figure*}

\begin{abstract}
Recent works have successfully extended large-scale text-to-image models to the video domain, producing promising results but at a high computational cost and requiring a large amount of video data. In this work, we introduce ConditionVideo, a training-free approach to text-to-video generation based on the provided condition, video, and input text, by leveraging the power of off-the-shelf text-to-image generation methods (\eg Stable Diffusion). ConditionVideo generates realistic dynamic videos from random noise or given scene videos.
Our method explicitly disentangles the motion representation into condition-guided and scenery motion components. To this end, the ConditionVideo model is designed with a UNet branch and a control branch. 
To improve temporal coherence, we introduce sparse bi-directional spatial-temporal attention (sBiST-Attn). The 3D control network extends the conventional 2D controlnet model, aiming to strengthen conditional generation accuracy by additionally leveraging the bi-directional frames in the temporal domain. 
Our method exhibits superior performance in terms of frame consistency, clip score, and conditional accuracy, outperforming compared methods. For the project website, see \url{https://pengbo807.github.io/conditionvideo-website/}
\end{abstract}

\section{Introduction}
Diffusion-based models \cite{song2021denoising,song2021scorebased,ho2020denoising,sohl2015deep} demonstrates impressive results in large-scale text-to-image (T2I) generation \cite{ramesh2022hierarchical,saharia2022photorealistic,gafni2022make,rombach2022high}. Much of the existing research proposes to utilize image generation models for video generation. Recent works \cite{singer2023makeavideo,blattmann2023align,DBLP:conf/iclr/Hong0ZL023} attempt to inflate the success of the image generation model to video generation by introducing temporal modules. While these methods reuse image generation models, they still require a massive amount of video data and training with significant amounts of computing power.
Tune-A-Video \cite{wu2022tune} extends Stable Diffusion \cite{rombach2022high}  with additional attention and a temporal module for video editing by tuning one given video. 
It significantly decreases the training workload, although an optimization process is still necessary. Text2Video \cite{khachatryan2023text2video} proposes training-free generation, however, the generated video fails to simulate natural background dynamics. Consequently, the question arises: \textit{How can we effectively utilize image generation models without any optimization process and embed controlling information as well as modeling dynamic backgrounds for video synthesis?}

We propose ConditionVideo, a training-free conditional-guided video generation method that utilizes off-the-shelf text-to-image generation models to generate realistic videos without any fine-tuning. Specifically, aiming at generating dynamic videos, our model disentangles the representation of motion in videos into two distinct components: conditional-guided motion and scenery motion, enabling the generation of realistic and temporally consistent frames. 
By leveraging this disentanglement, we propose a pipeline that consists of a UNet branch and a control branch,  with two separate noise vectors utilized in the sampling process. Each noise vector represents conditional-guided motion and scenery motion, respectively. To further enforce temporal consistency, 
we introduce sparse bi-directional spatial-temporal attention (sBiST-Attn) and a 3D control branch that leverages bi-directional adjacent frames in the temporal dimension to enhance conditional accuracy. 
Our ConditionVideo method outperforms the baseline methods in terms of frame consistency, conditional accuracy, and clip score. 

Our key contributions are as follows. (1) We propose ConditionVideo, a training-free video generation method that leverages off-the-shelf text-to-image generation models to generate conditional-guided videos with realistic dynamic backgrounds.
(2) Our method disentangles motion representation into conditional-guided and scenery motion components via a pipeline that includes a U-Net branch and a conditional-control branch.
(3) We introduce sparse bi-directional spatial-temporal attention (sBiST-Attn) and a 3D conditional-control branch to improve conditional accuracy and temporal consistency.

\section{Related Work}
\subsection{Diffusion Models}
Image diffusion models have achieved significant success in the field of generation \cite{ho2020denoising,song2021denoising,song2021scorebased}, surpassing numerous generative models that were once considered state-of-the-art \cite{dhariwal2021diffusion,kingma2021variational}. With the assistance of large language models \cite{radford2021learning,raffel2020exploring}, current research can generate videos from text, contributing to the prosperous of image generation \cite{ramesh2022hierarchical,rombach2022high}.

Recent works in video generation \cite{esser2023structure,DBLP:conf/nips/HoSGC0F22,wu2022tune,wu2021godiva,wu2022nuwa,DBLP:conf/iclr/Hong0ZL023,lavie,leo} aim to emulate the success of image diffusion models. Video Diffusion Models \cite{DBLP:conf/nips/HoSGC0F22} extends the UNet \cite{ronneberger2015u} to 3D and incorporates factorized spacetime attention \cite{bertasius2021space}. Imagen Video \cite{saharia2022photorealistic} scales this process up and achieves superior resolution. However, both approaches involve training from scratch, which is both costly and time-consuming. Alternative methods explore leveraging pre-trained text-to-image models. Make-A-Video \cite{singer2023makeavideo} facilitates text-to-video generation through an expanded unCLIP framework. Tune-A-Video \cite{wu2022tune} employs a one-shot tuning pipeline to generate edited videos from input guided by text. However, these techniques still necessitate an optimization process. 
Compared to these video generation methods, our training-free method can yield high-quality results more efficiently and effectively. 
 
\subsection{Conditioning Generation}

Recently, diffusion-based conditional video generation research has begun to emerge, gradually surpassing GAN-based methods \cite{mirza2014conditional,wang2018vid2vid,chan2019everybody,wang2019fewshotvid2vid,liu2019liquid,siarohin2019first,zhou2022cross,imaginator,g3an,lia}.
For the diffusion model-based image generation methods, a lot of works \cite{mou2023t2i,zhang2023adding} aim to enhance controllability through the integration of additional annotations.
ControlNet \cite{zhang2023adding} duplicates and fixes the original weight of the large pre-trained T2I model. Utilizing the cloned weight, ControlNet trains a conditional branch for task-specific image control.

Recent developments in the field of diffusion-based conditional video generation have been remarkable, branching into two main streams: text-driven video editing, as demonstrated by \cite{molad2023dreamix,esser2023structure,DBLP:journals/corr/abs-2303-12688,liu2023video,wang2023zero,qi2023fatezero,hu2023videocontrolnet}, and innovative video creation, featured in works like \cite{ma2023follow,khachatryan2023text2video,hu2023videocontrolnet,DBLP:journals/corr/abs-2305-13840,zhang2023controlvideo}. Our work is part of this exciting second stream.

In the realm of video generation, while systems like Follow-Your-Pose \cite{ma2023follow} and Control-A-Video \cite{DBLP:journals/corr/abs-2305-13840} are built upon an extensive training process, methods such as Text2Video-Zero \cite{khachatryan2023text2video} and ControlVideo \cite{zhang2023controlvideo} align more closely with our approach. A common challenge among these methods, however, is their limited capability in generating dynamic and vibrant backgrounds, a hurdle our methodology overcomes with our unique application of dynamic scene referencing. 
\begin{figure*}[t!]
  \centering
  \includegraphics[width=1\textwidth]{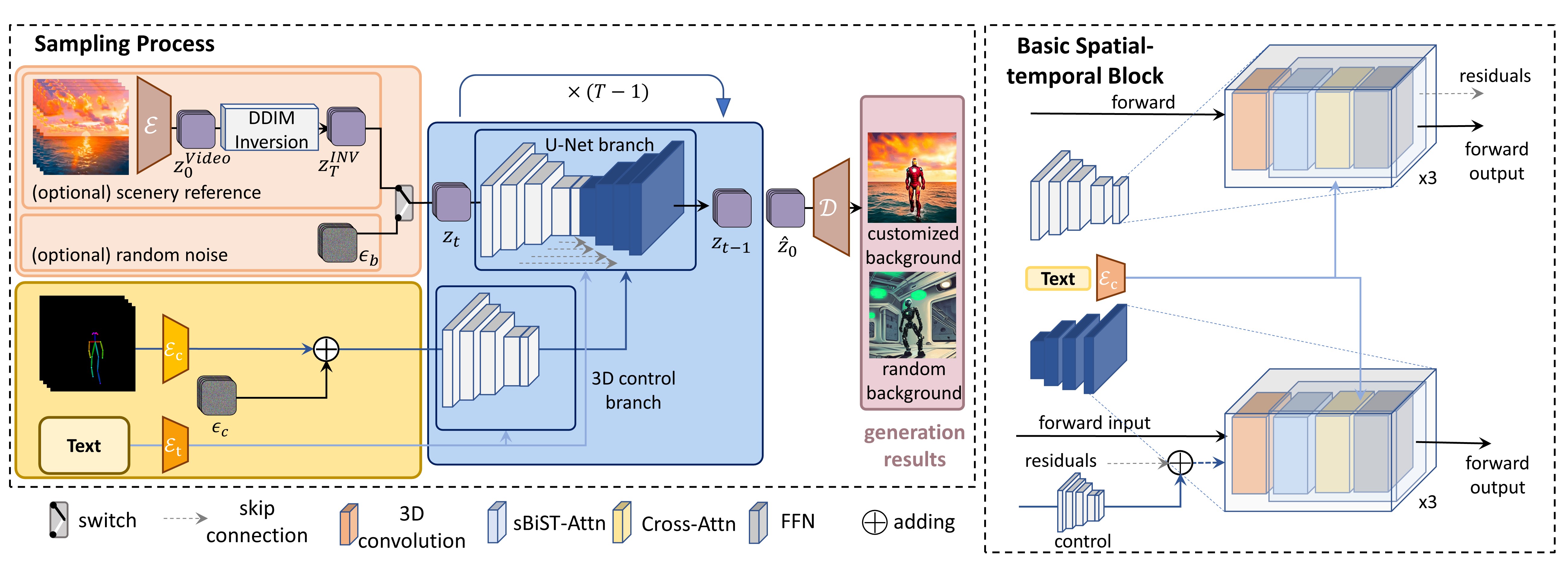}
  \caption{Illustration of our proposed training-free pipeline. (Left) Our framework consists of a UNet branch and a 3D control branch.  The UNet branch receives either the inverted reference video $z_T^{INV}$ or image-level noise $\epsilon_b$ for background generation. The 3D control branch receives an encoded condition for foreground generation. Text description is fed into both branches. (Right) Illustration of our basic spatial-temporal block. We employ our proposed sBiST-Attn module into the basic block between the 3D convolution block and the cross-attention block. The detail of sBiST-Attn module is shown in Fig. \ref{fig:attention}}
  \label{fig:pipeline}
\end{figure*}
\section{Preliminaries}
\paragraph{Stable Diffusion.}
\label{sec:stable-diffusion}
Stable Diffusion employs an autoencoder \cite{van2017neural} to preprocess images. An image $x$ in RGB space is encoded into a latent form by encoder $\mathcal{E}$ and then decoded back to RGB space by decoder $\mathcal{D}$. The diffusion process operates with the encoded latent $z=\mathcal{E}(x)$.

For the diffusion forward process, Gaussian noise is iteratively added to latent $z_0$ over $T$ iterations \cite{ho2020denoising}:
\begin{equation}
\begin{split}
q\left(z_t \mid z_{t-1}\right) &= \mathcal{N}\left(z_t ; \sqrt{1-\beta_t} z_{t-1}, \beta_t I\right), \\
t &= 1, 2, \ldots, T,
\end{split}
\end{equation}
where $q\left(z_t \mid z_{t-1}\right)$ denotes the conditional density function and $\beta$ is given.

The backward process is accomplished by a well-trained Stable Diffusion model that incrementally denoises the latent variable $\hat{z_0}$ from the noise $z_T$. Typically, the T2I diffusion model leverages a UNet architecture, with text conditions being integrated as supplementary information. The trained diffusion model can also conduct a deterministic forward process, which can be restored back to the original $z_0$. This deterministic forward process is referred to as DDIM inversion \cite{song2021denoising,dhariwal2021diffusion}.
We will refer to $z_T$ as the noisy latent code and $z_0$ as the original latent in the subsequent section. Unless otherwise specified, the frames and videos discussed henceforth refer to those in latent space.

\paragraph{ControlNet.}
ControlNet \cite{zhang2023adding} enhances pre-trained large-scale diffusion models by introducing extra input conditions. These inputs are processed by a specially designed conditioning control branch, which originates from a clone of the encoding and middle blocks of the T2I diffusion model and is subsequently trained on task-specific datasets. The output from this control branch is added to the skip connections and the middle block of the T2I model's UNet architecture. 

\section{Methods}
\label{sec:pipeline}
ConditionVideo leverages guided annotation, denoted as $Condtion$, and optional reference scenery, denoted as $Video$, to generate realistic videos. We start by introducing our training-free pipeline in Sec. \ref{sec:pipeline}, followed by our method for modeling motion in Sec. \ref{sec:motion-model}. In Sec. \ref{sec:sb-attn}, we present our sparse bi-directional spatial-temporal attention (sBiST-Attn) mechanism. Finally, a detailed explanation of our proposed 3D control branch is provided in Sec. \ref{sec:pose-guiding branch}.

\subsection{Training-Free Sampling Pipeline}

Fig. \ref{fig:pipeline} depicts our proposed training-free sampling pipeline. 
 Inheriting the autoencoder $\mathcal{D}(\mathcal{E}(\cdot))$ from the pre-trained image diffusion model (Sec. \ref{sec:stable-diffusion}), we conduct video transformation between RGB space and latent space frame by frame. Our ConditionVideo model contains two branches: a UNet branch and a 3D control branch. A text description is fed into both branches. Depending on the user's preference for customized or random background, the UNet branch accepts either the inverted code $z_T^{INV}$ of the reference background video or the random noise $\epsilon_b$. The condition is fed into the 3D control branch after being added with random noise $\epsilon_c$. We will further describe this disentanglement input mechanism and random noise $\epsilon_b$, $\epsilon_c$ in Sec. \ref{sec:motion-model}.

Our branch uses the original weight of ControlNet \cite{zhang2023adding}. As illustrated on the right side of Fig. \ref{fig:pipeline}, we modify the basic spatial-temporal blocks of these two branches from the conditional T2I model by transforming 2D convolution into 3D with 1$\times$3$\times$3 kernel and replacing the self-attention module with our proposed sBiST-Attn module (Sec. \ref{sec:sb-attn}). We keep other input-output mechanisms the same as before.
\subsection{Strategy for Motion Representation}
\label{sec:motion-model}
\paragraph{Disentanglement for Latent Motion Representation} 
In conventional diffusion models for generation (\eg ControlNet), the noise vector $\epsilon$ is sampled
from an i.i.d. Gaussian distribution $\epsilon \sim \mathcal{N}(0, I)$ and then shared by both the control branch and UNet branch. However, if we follow the original mechanism and let the inverse background video's latent code to shared by two branches, we observe that the background generation results will be blurred (Experiments are
shown in Appx. B.).  This is because using the same latent to generate both the foreground and the background presumes that the foreground character has a strong relationship with the background. 
Motivated by this observation, we explicitly disentangle the video motion presentation into two components: the motion of the background and the motion of the foreground. The background motion is generated by the UNet branch whose latent code is presented as background noise $\epsilon_b \sim \mathcal{N}(0, I)$. The foreground motion is represented by the given conditional annotations while the appearance representation of the foreground is generated from the noise $\epsilon_c\sim \mathcal{N}(0, I)$. 

\paragraph{Strategy for Temporal Consistency Motion Representation}
To attain temporal consistency across consecutively generated frames, We investigated selected noise patterns that facilitate the creation of cohesive videos. Consistency in foreground generation can be established by ensuring that the control branch produces accurate conditional controls. Consequently, we propose utilizing our control branch input for this purpose: $C_{cond} = \epsilon_c + \mathcal{E}_c(Condition),\epsilon_{c_i} \in \epsilon_c,\epsilon_{c_i} \sim \mathcal{N}(0, I)\subseteq \mathbb{R}^{H \times W \times C}, \forall i,j=1,...,F, \quad \epsilon_{ci}=\epsilon_{cj},$
where $H$, $W$, and $C$ denote the height, width, and channel of the latent $z_t$, $F$ represents the total frame number, $C_{cond}$ denotes the encoded conditional vector which will be fed into the control branch and $\mathcal{E}_c$ denotes the conditional encoder. Additionally, it's important to observe that $\epsilon_{c_i}$ corresponds to a single frame of noise derived from the video-level noise denoted as $\epsilon_{c}$. The same relationship applies to $\epsilon_{b_i}$ and $\epsilon_{b}$ as well.
\begin{algorithm}[htb]
\caption{Sampling Algorithm}
\label{Algotirhm}
\textbf{Input}: $Condition$, $Text$, $Video$(Optional)\\
\textbf{Parameter}: $T$\\
\textbf{Output}: $\hat{X}_0$:generated video
\begin{algorithmic}[1] 
    \IF{$Video$ is not None}           
        \STATE $z_0^{Video}\gets \mathcal{E}(Video)\quad$ //encode video
        \STATE $z_T^{INV} \gets \text{DDIM\_Inversion}(z_0^{Video},T,\text{UNetBranch})$
        \STATE $z_T \gets z_T^{INV}\quad$   //customize background
    \ELSE
        \STATE $z_T \gets \epsilon_b,\quad$         //random background
    \ENDIF
    \STATE $C_{cond} \gets \epsilon_c + \mathcal{E}_c(Condition)\quad$ //encode condition
    \STATE $C_{text} \gets \mathcal{E}_t(Text)$ //encode input prompt
    \FOR{$t=T...1$}
        \STATE $c_t \gets \text{ConrtolBranch}(C_{cond},t,C_{text})$
        \STATE $\hat{z}_{t-1}\gets \text{DDIM\_Backward}(z_t,t,C_{text},c_t,$\\$\text{UNetBranch})$ 
    \ENDFOR
    \STATE $\hat{X}_0 \gets \mathcal{D}(\hat{z}_0)$
    \STATE \textbf{return} $\hat{X}_0$
\end{algorithmic}
\end{algorithm}
When generating backgrounds, there are two approaches we could take. The first is to create the background using background noise $\epsilon_b$:
$\epsilon_{b_i} \in \epsilon_b,~~
    \epsilon_{b_i} \sim \mathcal{N}(0, I)\subseteq \mathbb{R}^{H \times W \times C} 
    \\\epsilon_{bi}=\epsilon_{bj} ,\quad \forall i,j=1,...,F.$
The second approach is to generate the background from an inverted latent code, $z_T^{INV}$, of the reference scenery video. Notably, we observed that the dynamic motion correlation present in original video is retained when it undergoes DDIM inversion. So we utilize this latent motion correlation to generate background videos. 

During the sampling process, in the first forward step $t=T$, we feed the background latent code $z_T^{INV}$ or $\epsilon_b$ into the UNet branch and the condition $C_{cond}$ into our 3D control branch. Then, during the subsequent reverse steps $t=T-1, .., 0$, we feed the denoised latent $z_t$ into the UNet branch while still using $C_{cond}$ for 3D control branch input. The details of the sampling algorithm are shown in Alg. \ref{Algotirhm}

\subsection{Sparse Bi-directional Spatial-Temporal Attention (sBiST-Attn)}
\label{sec:sb-attn}

\begin{figure}[h]
  \centering
  \includegraphics[width=0.45\textwidth]{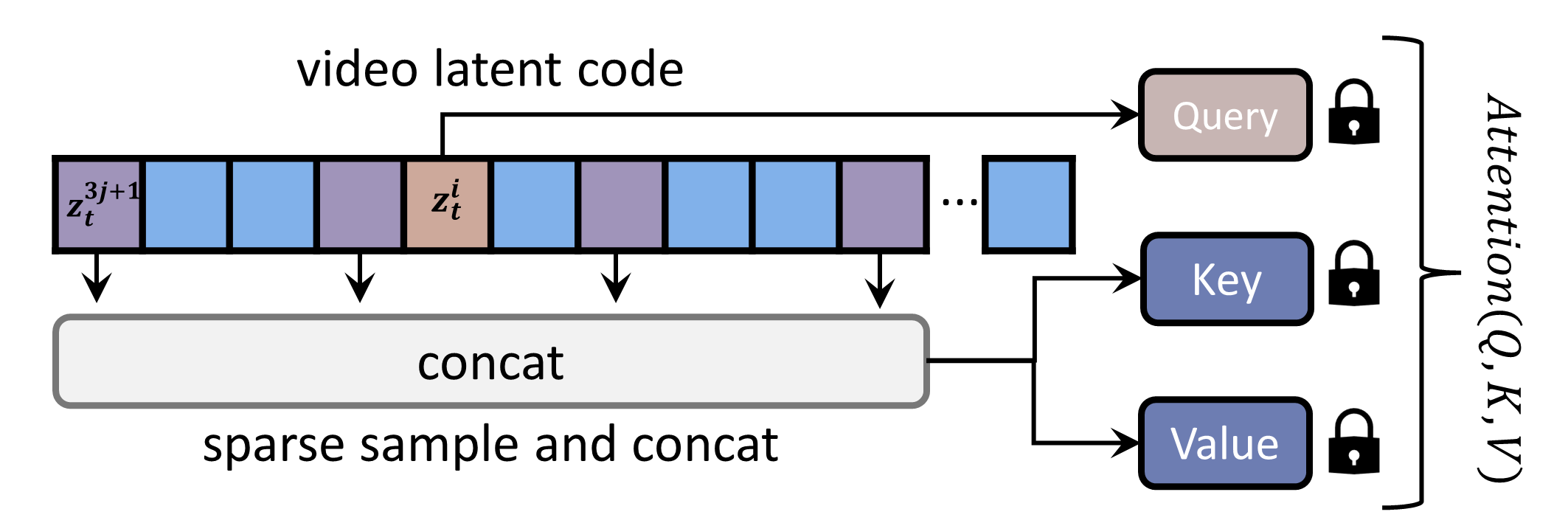}
  \caption{Illustration of ConditionVideo's sBiST-Attn. The purple blocks signify the frame we've selected for concatenation, which can be computed for key and value. The pink block represents the current block from which we'll calculate the query. The blue blocks correspond to the other frames within the video sequence. Latent features of frame $z_t^i$, bi-directional frames $z_t^{3j+1}, ~ j=0,...,\lfloor (F-1)/3 \rfloor$ are projected to query $Q$, key $K$ and value $V$. Then the attention-weighted sum is computed based on key, query, and value. The parameters are the same as the ones in the self-attention module of the pre-trained image model.}
  \label{fig:attention}
\end{figure}
Taking into account both temporal coherence and computational complexity, we propose a sparse bi-directional spatial-temporal attention (sBiST-Attn) mechanism, as depicted in Fig. \ref{fig:attention}. For video latent $z_t^i, ~i=1,...,F$, the attention matrix is computed between frame $z_t^i$ and its bi-directional frames, sampled with a gap of 3. This interval was chosen after weighing frame consistency and computational cost (see Appx. C.1). For each $z_t^i$ in $z_t$, we derive the query feature from its frame $z_t^i$. The key and value features are derived from the bi-directional frames $z_t^{3j+1}, ~ j=0,...,\lfloor (F-1)/3 \rfloor$. Mathematically, our sBiST-Attn can be expressed as:

 \begin{equation}
\left\{
\begin{aligned}
&\mathrm{Attention}(Q, K, V) = \mathrm{Softmax}\left(\frac{Q K^T}{\sqrt{d}}\right) \cdot V \\
&Q = W^Q z_t^i,  K = W^K z_t^{[3j+1]},  V = W^V z_t^{[3j+1]}, \\
&j = 0, 1, \ldots, \lfloor (F-1)/3 \rfloor
\end{aligned}
\right.
\end{equation}
where [·] denotes the concatenation operation, and $W^Q, W^K, W^V$ are the weighted matrices that are identical to those used in the self-attention layers of the image generation model.

\subsection{3D Control Branch}
\label{sec:pose-guiding branch}
Frame-wise conditional guidance is generally effective, but there may be instances when the network doesn't correctly interpret the guide, resulting in an inconsistent conditional output. Given the continuous nature of condition movements, ConditionVideo propose enhancing conditional alignment by referencing neighboring frames. If a frame isn't properly aligned due to weak control, other correctly aligned frames can provide more substantial conditional alignment information. In light of this, we design our control branch to operate temporally, where we choose to replace the self-attention module with the sBiST-Attn module and inflate 2D convolution to 3D. The replacing attention module can consider both previous and subsequent frames, thereby bolstering our control effectiveness.

\section{Experiments}
\subsection{Implementation Details}
We implement our model based on the pre-trained weights of ControlNet  \cite{zhang2023adding} and Stable Diffusion \cite{rombach2022high} 1.5.
We generate 24 frames with a resolution of 512 × 512 pixels for each video. 
During inference, we use the same sampling setting as Tune-A-Video \cite{wu2022tune}. More details can be found in Appx. D at \url{https://arxiv.org/abs/2310.07697}.
\subsection{Main results}
In Fig. \ref{fig:result}, we display the success of our training-free video generation technique. The generated results from ConditionVideo, depicted in Fig. \ref{fig:result} (a), imitate moving scenery videos and show realistic waves as well as generate the correct character movement based on posture. Notably, the style of the backgrounds is distinct from the original guiding videos, while the motion of the backgrounds is the same. Furthermore, our model can generate consistent backgrounds when sampling $\epsilon_b$ from Gaussian noise based on conditional information, as shown in Fig.\ref{fig:result} (b),(c),(d). These videos showcase high temporal consistency and rich graphical content.
\subsection{Comparison}

\subsubsection{Compared Methods}
We compare our method with Tune-A-Video \cite{wu2022tune}, ControlNet \cite{zhang2023adding}, and Text2Video-Zero \cite{khachatryan2023text2video}. For Tune-A-Video, we first fine-tune the model on the video from which the condition was extracted, and then sample from the corresponding noise latent code of the condition video.
\begin{figure}[htb]
  \centering
\includegraphics[width=0.45\textwidth]{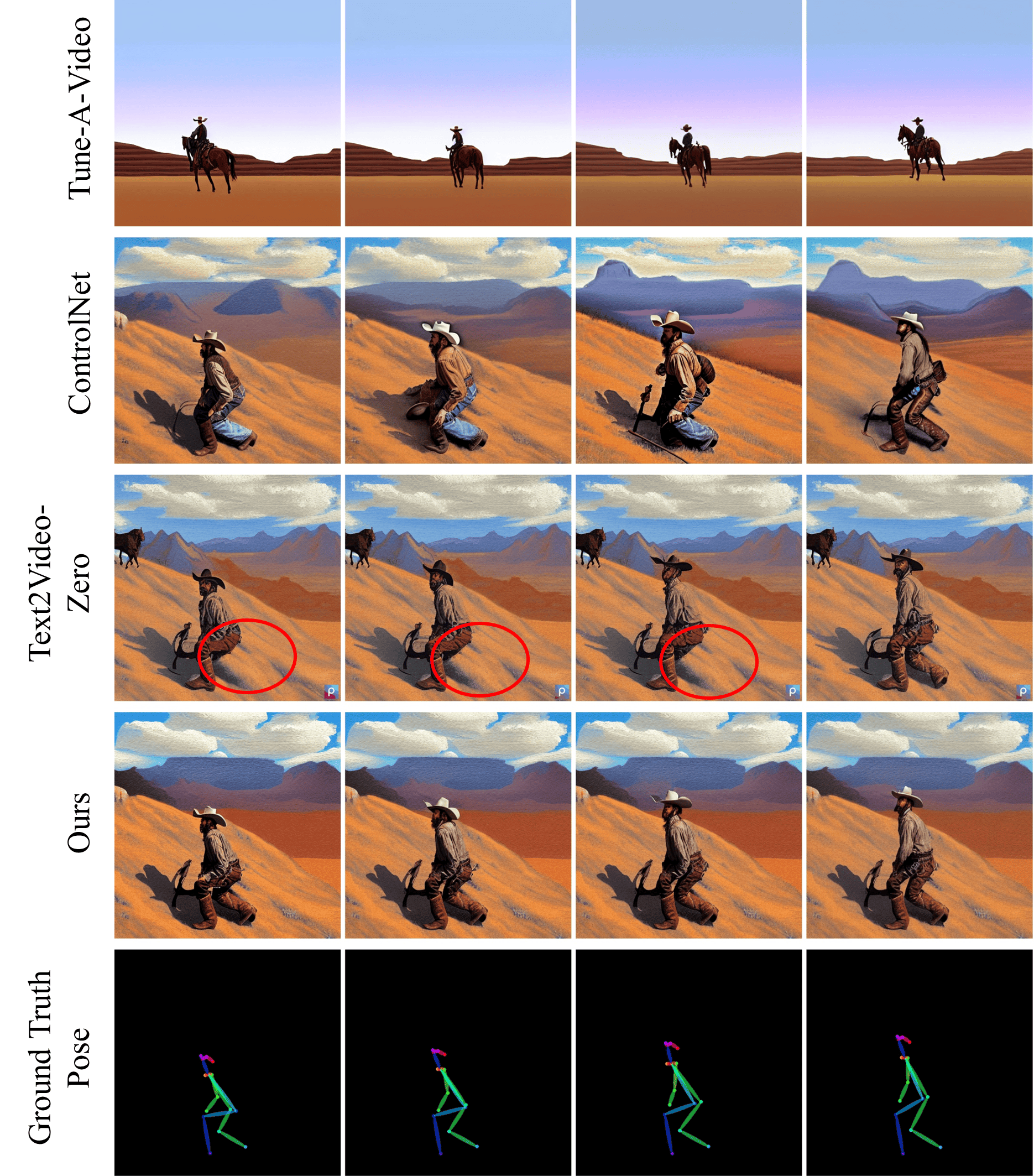}
  \caption{Qualitative comparison condition on the pose. \textit{``The Cowboy, on a rugged mountain range, Western painting style".} Our result outperforms in both temporal consistency and pose accuracy, while others have difficulty in maintaining either one or both of the qualities.}
  \label{fig:quality-compare-pose}
\end{figure}
\subsubsection{Qualitative Comparison}
\begin{figure}[htb]
  \centering
  \includegraphics[width=0.45\textwidth]{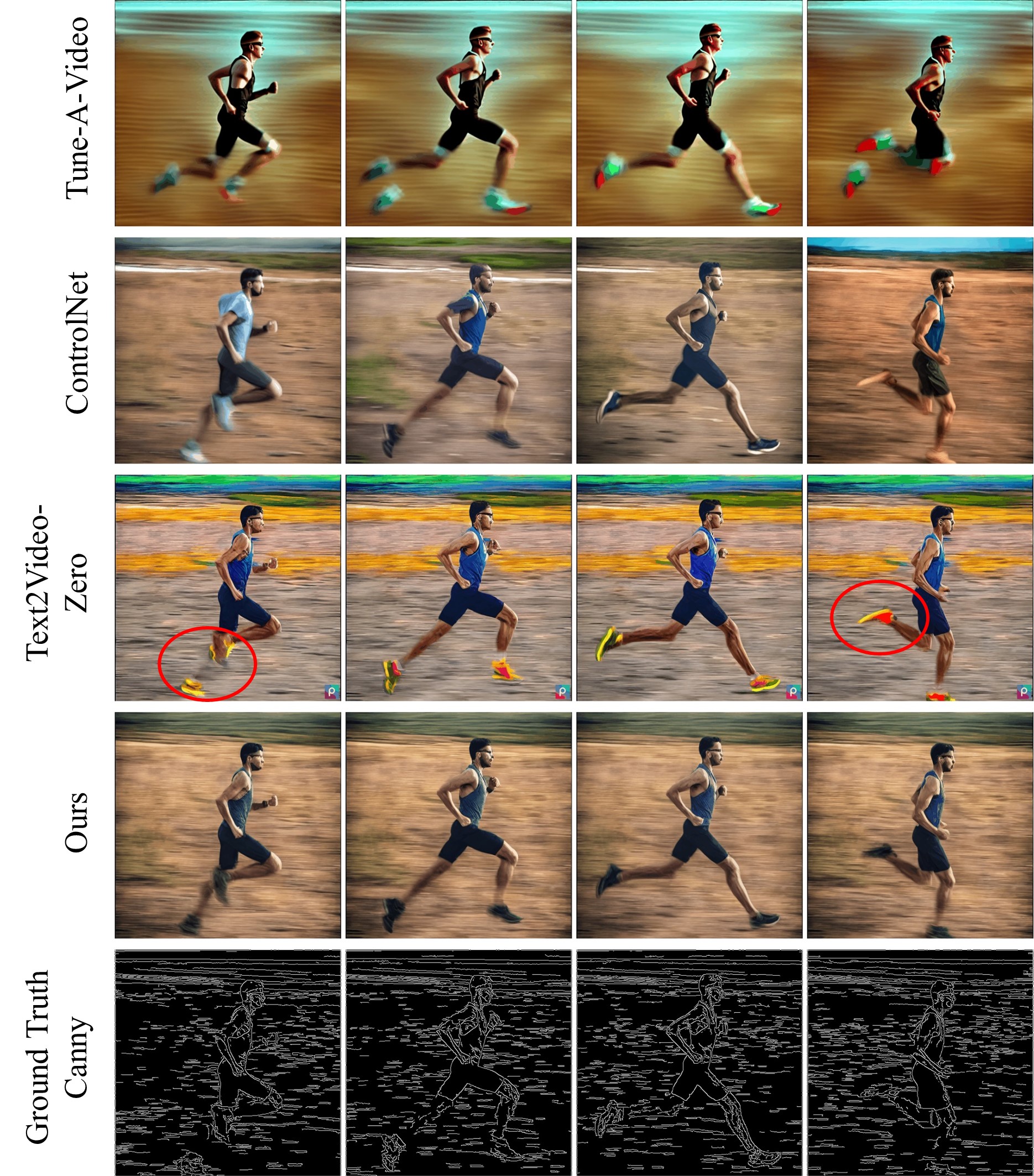}
  \caption{Qualitative comparison condition on canny. \textit{``A man is runnin''.} Tune-A-Video experiences difficulties with canny-alignment, while ControlNet struggles to maintain temporal consistency. Though Text2Video surpasses these first two approaches, it inaccurately produces parts of the legs that don't align with the actual human body structure, and the colors of the shoes it generates are inconsistent.}
  \label{fig:quality-compare-canny}
\end{figure}

\begin{figure}[htb]
  \centering
  \includegraphics[width=0.45\textwidth]{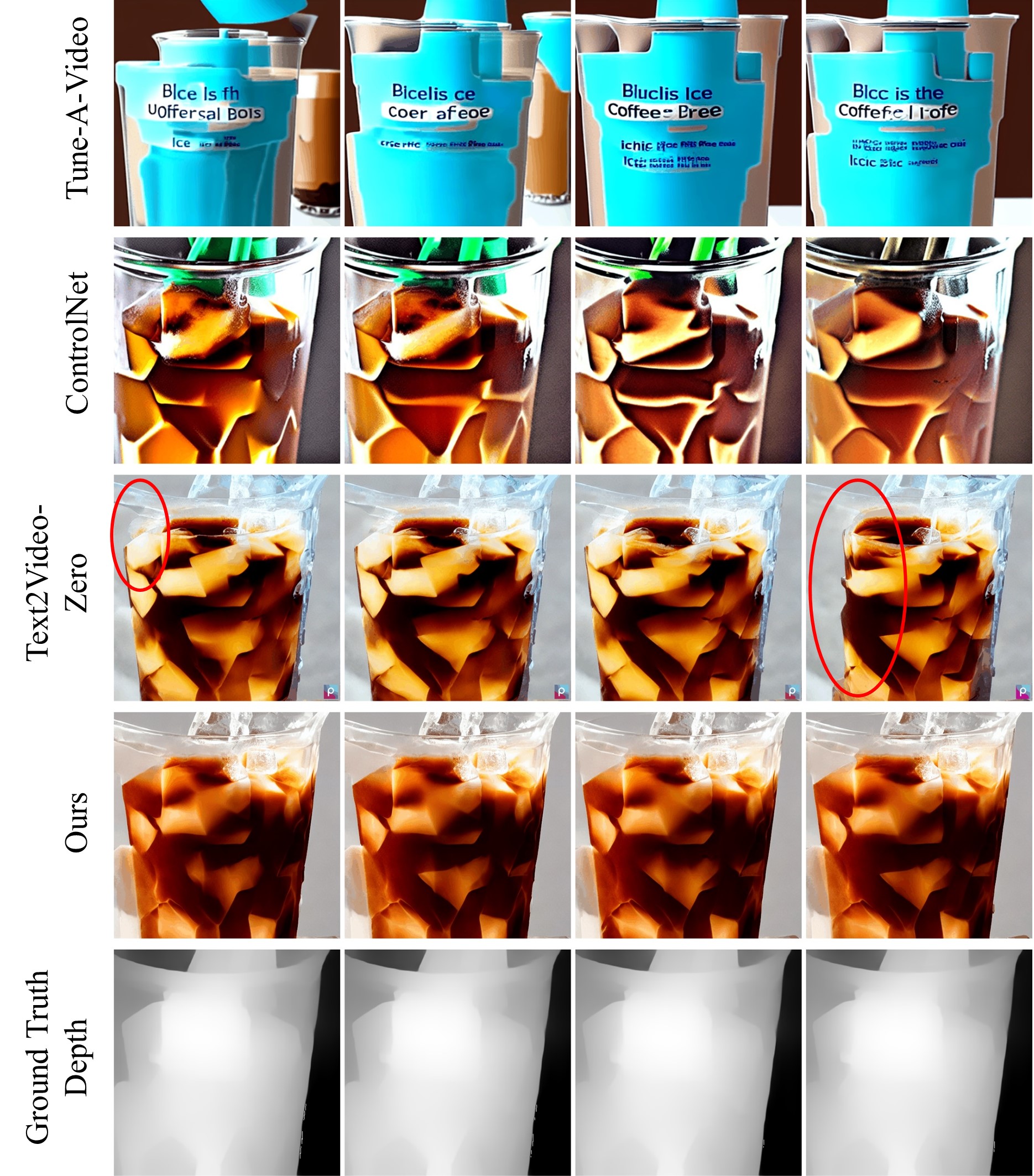}
  \caption{Qualitative comparison condition on depth. \textit{``ice coffee".} All three methods used for comparison have the problem of changing the appearance of the object when the viewpoint is switched, and only our method ensures the consistency of the appearance before and after.}
  \label{fig:quality-compare-depth}
\end{figure}

Our visual comparison conditioning on pose, canny, and depth information is presented in Fig. \ref{fig:quality-compare-pose}, \ref{fig:quality-compare-canny}, and \ref{fig:quality-compare-depth}. Tune-A-Video struggles to align well with our given condition and text description. ControlNet demonstrates improvement in condition-alignment accuracy but suffers from a lack of temporal consistency. Despite the capability of Text2Video to produce videos of exceptional quality, there are still some minor imperfections that we have identified and indicated using a red circle in the figure. Our model surpasses all others, showcasing outstanding condition-alignment quality and frame consistency.

\subsubsection{Quantitative Comparison}
\label{sec:quantitative}
\begin{table}[thb]
  \centering
  \begin{tabular}{lllll}
    \toprule
    Method               & FC(\%)     & CS              & PA (\%) \\
    \midrule
    Tune-A-Video  & 95.84& 30.74 &26.13 \\
    ControlNet & 94.22& 32.97 &79.51\\
    Text2Video-Zero& 98.82& 32.84 &78.50 \\
    Ours&\textbf{99.02}&\textbf{33.03}&\textbf{83.12}\\
    \bottomrule
  \end{tabular}
  \caption{Quantitative comparisons condition on pose. FC, CS, PA represent \textit{frame consistency}, \textit{clip score} and \textit{pose-accuracy}, respectively}
  \label{quantitative-table-pose}
\end{table}
\begin{table}[thb]
  \centering
  \begin{tabular}{lllll}
    \toprule
    Method                       & Condition & FC(\%)     & CS    \\
    \midrule
    Tune-A-Video                & - & 95.84& 30.74\\
    \midrule
    ControlNet   & Canny            & 90.53& 29.65 \\
    Text2Video-Zero & Canny& 97.44& 28.76  \\
    Ours                    & Canny&\textbf{97.64}&\textbf{29.76}\\
    \midrule
    ControlNet & Depth            & 90.63& 30.16 \\
    Text2Video-Zero & Depth& 97.46& 29.38  \\
    Ours                    & Depth&\textbf{97.65}&\textbf{30.54}\\
    \midrule
    ControlNet & Segment            & 91.87& 31.85 \\
    Ours       & Segment&\textbf{98.13}&\textbf{32.09}\\
    \bottomrule
  \end{tabular}
  \caption{Quantitative comparisons condition on canny, depth and segment.}
  \label{quantitative-table-othercond}
\end{table}
We evaluate all the methods using three metrics: \textit{frame consistency} \cite{esser2023structure,wang2023zero,radford2021learning}, \textit{clip score} \cite{ho2022imagen,hessel-etal-2021-clipscore,park2021benchmark}, and \textit{pose accuracy} \cite{ma2023follow}. As other conditions are hard to evaluate, we use pose accuracy for conditional consistency only. The results on different conditions are shown in Tab. \ref{quantitative-table-pose} and \ref{quantitative-table-othercond}. We achieve the highest frame consistency, and clip score in all conditions, indicating that our method exhibits the best time consistency and text alignment. We also have the best pose-video alignment among the other three techniques. The conditions are randomly generated from a group of 120 different videos. For more information please see Appx. D.2.

\subsection{Ablation Study}
\begin{figure}[t!]
  \centering
  \includegraphics[width=0.45\textwidth]{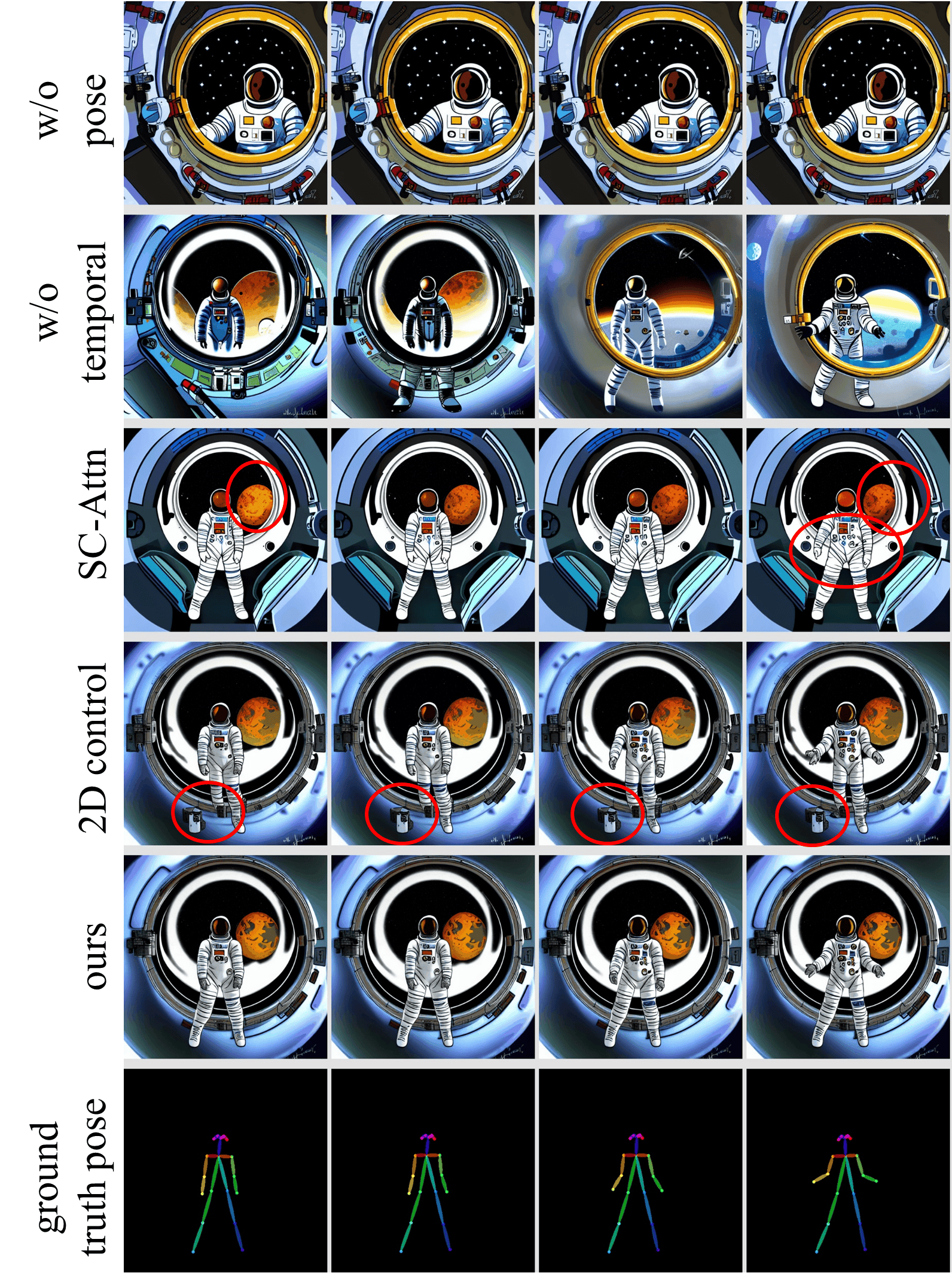}
  \caption{Ablations of each component, generated from image-level noise. \textit{``The astronaut, in a spacewalk, sci-fi digital art style''}.
1st row displays the generation result without pose conditioning. 2nd and 3rd rows show the results after replacing our sBiST-Attn with self-Attn and SC-Attn \cite{wu2022tune}. 4th row presents the result with the 2D condition-control branch.}
  \label{fig:ablation-noise}
\end{figure}
\begin{table}[t]
  \centering
  \begin{tabular}{lllll}
    \toprule
    Method                & FC(\%) & Time \\
    \midrule
    w/o Temp-Attn         & 94.22  & \textbf{31s}\\
    S-C Attn&             98.77    & 43s\\
    sBiST-Attn&           99.02    & 1m30s\\
    Full-Attn&            \textbf{99.03} & 3m37s\\
    \bottomrule
  \end{tabular}
  \caption{Ablations on temporal module. Time represents the duration required to generate a 24-frame video with a size of 512x512.}
  \label{temp-module-ablation-table}
\end{table}

We conduct an ablation study on the pose condition, temporal module, and 3D control branch. The qualitative result is visualized in Fig. \ref{fig:ablation-noise}. In our research, we modify each element individually for comparative analysis, ensuring that all other settings remain constant.

\paragraph{Ablation on Pose Condition}
We evaluate performance with and without using pose, as shown in Fig. \ref{fig:ablation-noise}. Without pose conditioning, the video is fixed as an image, while the use of pose control allows for the generation of videos with certain temporal semantic information.

\paragraph{Ablation on Temporal Module}


Training-free video generation heavily relies on effective spatial-temporal modeling. To evaluate the efficacy of our temporal attention module, We remove our sBiST-attention mechanism and replace it with a non-temporal self-attention mechanism, a Sparse-Causal attention mechanism \cite{wu2022tune} and a dense attention mechanism \cite{wang2023zero} which attends to all frames for key and value.

The results are presented in Tab.~\ref{temp-module-ablation-table}. A comparison of temporal and non-temporal attention underlines the importance of temporal modeling for generating time-consistent videos. By comparing our method with Sparse Causal attention, we demonstrate the effectiveness of ConditionVideo's sBiST attention module, proving that incorporating information from bi-directional frames improves performance compared to using only previous frames. Furthermore, we observe almost no difference in frame consistency between our method and dense attention, despite the latter requiring more than double our generation duration.
\paragraph{Ablations on 3D Control Branch}
\begin{table}[t]
  \centering
  \begin{tabular}{llll}
    \toprule
    Method                  & FC(\%)         & CS  (\%)       & PA (\%)\\
    \midrule
    2D control          & \textbf{99.03} & \textbf{33.11}           & 81.26 \\
    3D control          & 99.02                         & 33.03     &\textbf{83.12}  \\
    \bottomrule
  \end{tabular}
  \caption{Ablation on 3D control branch. FC, CS, PA represent frame consistency, clip score, and pose-accuracy, respectively.}
  \label{ControlNet-ablation-table}
\end{table}

We compare our 3D control branch with a 2D version that processes conditions frame-by-frame. For the 2D branch, we utilize the original ControlNet conditional branch. Both control branches are evaluated in terms of frame consistency, clip score, and pose accuracy. Results in Tab. \ref{ControlNet-ablation-table} show that our 3D control branch outperforms the 2D control branch in pose accuracy while maintaining similar frame consistency and clip scores. This proves that additional consideration of bi-directional frames enhances pose control.


\section{Discussion and Conclusion}
In this paper, we propose ConditionVideo, a training-free method for generating videos with vivid motion. This technique leverages a unique motion representation, informed by background video and conditional data, and utilizes our sBiST-Attn mechanism and 3D control branch to enhance frame consistency and condition alignment. Our experiments show that ConditionVideo can produce high-quality videos, marking a significant step forward in video generation and AI-driven content creation.

During our experiments, we find that our method is capable of generating long videos. Moreover, this approach is compatible with the hierarchical sampler from ControlVideo \cite{zhang2023controlvideo}, which is used for generating long videos. 
Despite the effectiveness of condition-based and temporal attention in maintaining video coherence, challenges such as flickering in videos with sparse conditions like pose data were noted. To address this issue, a potential solution would involve incorporating more densely sampled control inputs and additional temporal-related structures.

\section*{Acknowledgements}
This work was jointly supported by the National Key R\&D Program of China(NO.2022ZD0160100) and the National Natural Science Foundation of China under Grant No. 62102150.

\bibliography{aaai24}

\end{document}